\documentclass[letterpaper, 10 pt, conference]{ieeeconf}  

\IEEEoverridecommandlockouts                              

\title{\LARGE \bf
SweepNet: Wide-baseline Omnidirectional Depth Estimation
}

\author{Changhee Won, Jongbin Ryu and Jongwoo Lim$^*$\vspace{3pt}\\
{\tt\small \{chwon, jongbinryu, jlim\}@hanyang.ac.kr}\\
{\small Department of Computer Science, Hanyang University, Seoul, Korea.}
}
\usepackage{gensymb}
\usepackage{graphicx}
\usepackage[font=small,labelfont=bf,singlelinecheck=false]{caption}
\usepackage[subrefformat=parens]{subcaption}
\usepackage{kotex}
\usepackage{bm}
\usepackage{amsmath}
\usepackage{amssymb}
\let\labelindent\relax
\usepackage{enumitem}
\usepackage{subcaption}
\usepackage{color}
\usepackage{booktabs}
\usepackage{multirow}
\usepackage{siunitx}
\usepackage{threeparttable}

\makeatletter
\let\NAT@parse\undefined
\makeatother

\usepackage{cite}
\usepackage{hyperref}
\hypersetup{colorlinks,linktocpage=true}

\newcommand{\etal}[1]{#1~\textit{et al.}}

\newcommand{\subsec}[1]{\vspace{4pt}{\setlength{\parindent}{0pt}\textbf{#1}~}}
\graphicspath{{./img/}}  

\usepackage{tikz}
\usepackage{textcomp}
\usepackage{hyperref}
\usepackage{lipsum}

\newcommand\copyrighttext{%
  \footnotesize \textcopyright 2019 IEEE. Personal use of this material is permitted.
  Permission from IEEE must be obtained for all other uses, in any current or future
  media, including reprinting/republishing this material for advertising or promotional
  purposes, creating new collective works, for resale or redistribution to servers or
  lists, or reuse of any copyrighted component of this work in other works.
  DOI: \href{https://doi.org/10.1109/ICRA.2019.8793823}{DOI No. 10.1109/ICRA.2019.8793823}}
\newcommand\copyrightnotice{%
\begin{tikzpicture}[remember picture,overlay]
\node[anchor=south,yshift=10pt] at (current page.south) {\fbox{\parbox{\dimexpr\textwidth-\fboxsep-\fboxrule\relax}{\copyrighttext}}};
\end{tikzpicture}%
}

\begin{document}

\maketitle
\copyrightnotice
\thispagestyle{empty}
\pagestyle{empty}

\begin{abstract}

Omnidirectional depth sensing has its advantage over the conventional stereo systems since it enables us to recognize the objects of interest in all directions without any blind regions.
In this paper, we propose a novel wide-baseline omnidirectional stereo algorithm which computes the dense depth estimate from the fisheye images using a deep convolutional neural network.
The capture system consists of multiple cameras mounted on a wide-baseline rig with ultra-wide field of view (FOV) lenses, and we present the calibration algorithm for the extrinsic parameters based on the bundle adjustment.
Instead of estimating depth maps from multiple sets of rectified images and stitching them, our approach directly generates one dense omnidirectional depth map with full 360\degree\ coverage at the rig global coordinate system.
To this end, the proposed neural network is designed to output the cost volume from the warped images in the sphere sweeping method, and the final depth map is estimated by taking the minimum cost indices of the aggregated cost volume by SGM.
For training the deep neural network and testing the entire system, realistic synthetic urban datasets are rendered using Blender.
The experiments using the synthetic and real-world datasets show that our algorithm outperforms the conventional depth estimation methods and generate highly accurate depth maps.

\end{abstract}

\section{INTRODUCTION}
\label{sec:introduction}
Estimating 3D geometry is an essential part of many robotic tasks such as navigation, recognition, manipulation, and planning.
Many sensor systems including LIDAR, structured-light 3D scanner, or stereo cameras have been developed and used to this end.
Among these methods the camera-based stereo systems have many benefits as they operate in a passive mode (not emitting any active signal), and they are compact, light-weight, and mechanically robust.
%
%
Moreover, thanks to the recent advances in GPU processors and deep learning algorithms, the camera-based methods become feasible and promising.

%
A conventional stereo setup uses two cameras looking in the same direction at a horizontal interval to estimate the disparity map.
However, in a real-world environment where obstacles exist around the robot, it is often necessary to estimate the omnidirectional depth map. 
The popular methods \cite{shimamura2000construction, wang2012stereo} use multiple stereo cameras, estimate the disparity maps from rectified image pairs, and then merge them into one panoramic image.
The reconstruction results are mostly favorable, but the size and cost of the system can be problematic in some cases.
Algorithmically the distortion in the boundary of the rectified images can cause incorrect depth estimate and the discontinuity in an overlapping area makes fusing multiple disparities difficult.
To reduce the number of cameras and the rig size, researchers have proposed two vertically-mounted cameras with wide FOV fisheye lenses, 360\degree\ catadioptric lenses or reflective mirrors to get a pair of omnidirectional images.
%
%
%
In this setup the rectified images and disparity maps are in low resolution, and the vertical epipolar lines make hard to estimate the depth of vertical structures.
Also when long-range sensing is needed, as in autonomous driving, the short baseline in the above systems can limit the effective sensing distance as it is proportional to the baseline between the cameras.

\begin{figure}
\centering
    
	\includegraphics[width=\linewidth]{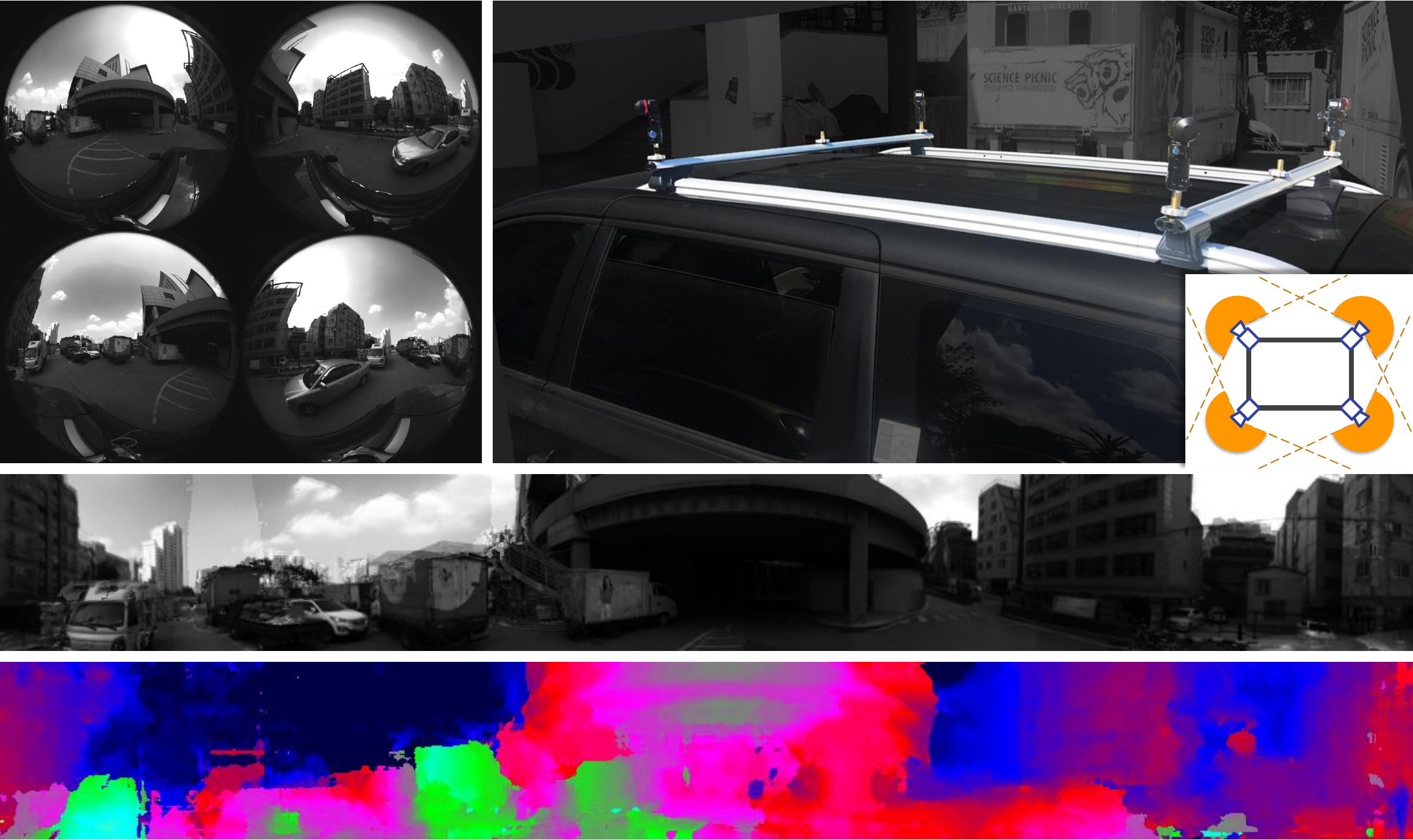}
    
	\caption{Top: an example of input images and our omnidirectional capture system installed on a minivan. Middle: reference panorama image obtained by using the inverse depth map. Bottom: estimated omnidirectional inverse depth map.}
    \label{fig:intro}
    \vspace{-15pt}
\end{figure}
   
In this paper, we propose a novel wide-baseline omnidirectional stereo vision system which uses only four cameras but estimates full and continuous omnidirectional depth map.
Each camera is equipped with a 220\degree\ FOV fisheye lens and facing the four cardinal directions as shown in Fig.~\ref{fig:intro}. 
The proposed system can generate an omnidirectional depth map of 360\degree\ horizontal FOV and up to 180\degree\ vertical FOV (full sphere). 
This new camera and lens configuration yields much larger overlapped region of view frustums which enables the robots or cars to sense nearby obstacles in narrow or crowded environments.
%
%
%
%

In multi-camera stereo systems, the plane sweep method~\cite{collins1996space} which sweeps parallel virtual planes, and projects the input images onto the planes to find stereo correspondences has been used.
\etal{Gallup}~\cite{gallup2007real} propose a more robust method that uses multiple sweeping directions, and it is further extended for the fisheye images \cite{hane2014real}. 
In this paper, we propose the spherical sweeping method similar to \etal{Im} \cite{im2016all}.
Instead of parallel planes, the concentric virtual spheres centered at the rig coordinate system are swept for the predefined range of inverse depths.
The input fisheye images are individually projected onto the spheres and the matching costs are computed from these projected spherical images.
In this way, continuous omnidirectional depth maps can be generated without artificially dividing views and stitching them later, and different camera configurations (numbers and positions) can be seamlessly handled.
%
%
%
%

In our setup, depending on ray directions, the minimum depth estimate ranges 0.8$\sim$1.7$\times$ the distance between cameras, which is extremely wide-baseline in the stereo literature.
%
%
The image patches at near distance suffer severe appearance changes due to large geometric and radiometric variation between views.
Since the conventional patch-based cost metrics use the local information, the generated cost volumes are noisy and often miss correct matches even after cost aggregation.
We observe that even the recent deep learning-based models \cite{zbontar2016stereo} suffer a similar problem.
Therefore, we propose a novel neural network-based approach that considers global context information in cost computation.
The proposed network takes the spherical images of two views and outputs a whole omnidirectional cost map.
The extensive experiments show that the proposed network outperforms the conventional local matching methods.

Our contributions are summarized as:
\begin{enumerate}[label=(\roman*)]
\setlength{\topsep}{0pt}
\setlength{\itemsep}{0pt}
\setlength{\partopsep}{0pt} 
\item 
We propose a novel omnidirectional wide-baseline stereo system which can estimate 360\degree\ dense depth maps up to very close distance.
Both the hardware configuration of a small number of cameras with ultra-wide FOV lenses and the software system for depth estimation are new, flexible, and effective in accomplishing the proposed goal.
%
%
%
%
%
%
\item 
We design a deep neural network for computing the matching cost map for a pair of spherical images.
The sphere sweeping at the rig coordinate system effectively normalizes the captured images into a uniform input to the network, and by using the whole spherical images, the network learns global contexts for more accurate stereo matching.
%
%
\item
The realistic synthetic urban datasets are rendered for training and testing the deep neural network.
With this datasets the proposed algorithm is compared with the previous algorithms by extensive quantitative evaluation.
Further, real-world datasets are collected to show the performance of the proposed system.
All datasets, as well as the trained network, will be made public when the paper is published.

\end{enumerate}
%
%

\section{RELATED WORK}
\label{sec:related}

\subsec{Omnidirectional Stereo}
%
There have been three major approaches for the omnidirectional stereo vision system: spinning a camera, using mirrors, and using wide FOV fisheye lenses. 
\etal{Kang} \cite{kang19973} and \etal{Peleg} \cite{peleg2001omnistereo} compute one panoramic depth map from multiple images captured while spinning an arm with a camera at the end.
Despite the advantage of using one camera, long capture time and the rotating arm make it difficult to be used outside the lab.
Using mirrors with a few cameras is popular in omnidirectional stereo systems \cite{bunschoten2003robust,yi2006omnidirectional,shimamura2000construction}. 
%
\etal{Geyer} \cite{geyer2003conformal} and \etal{Sch{\"o}nbein} \cite{schonbein2014omnidirectional} use two horizontally-mounted cameras with 360\degree\ FOV catadioptric lenses.
Although they can generate an omnidirectional depth map, there exist two blind spots along the epipole direction, and the depth estimates around them are unstable or missing.
%
%
Gao and Shen \cite{gao2017dual} propose a system with two vertically-mounted cameras with ultra-wide FOV fisheye lenses.
%
It performs omnidirectional depth estimation by projecting the input fisheye images into four virtual planes parallel to the baseline.
%
However, the disparity maps are in low resolution due to the limitation of sensor resolution and high distortion by the fisheye lenses, and the depth estimates of vertical structures parallel to the baseline are often unavailable.
%
%
Meanwhile, an omnidirectional motion stereo algorithm \etal{Im} \cite{im2016all} is presented - it computes a 360\degree\ depth map of the static scene from a short video clip captured by a moving omnidirectional camera.
The sphere sweeping method allows the images to be captured at any known poses, thus lifts the fixed configuration restriction.
While they address a motion stereo in a very short-baseline setup where appearance variations across views are minimal, we try to solve a more challenging extremely wide-baseline problem.

\subsec{Stereo Matching Cost}
According to \etal{Scharstein} \cite{scharstein2002taxonomy}, there are four steps in stereo depth estimation: initial matching cost computation, cost aggregation, disparity computation with optimization, and disparity refinement.
Among them, computing matching costs from the input images is the most demanding and difficult part.
%
%
Typical intensity-based matching costs include sum of absolute differences (SAD),
normalized cross-correlation (NCC), rank, or census transforms \cite{zabih1994non}. 
\etal{Hirschmuller} \cite{hirschmuller2007evaluation} compares and evaluates these matching cost functions. 

Instead of finding the minimum values of locally aggregated matching costs, the depth map can be computed by global optimization by graph cuts \cite{kolmogorov2001computing} or belief propagation \cite{klaus2006segment,bleyer2011patchmatch}, but they require high computational cost. 
%
%
Semi-global matching (SGM) \cite{hirschmuller2008stereo} is an efficient way of aggregating costs globally using dynamic programming.

Recently, due to the large-scale stereo datasets with ground truth depths \cite{geiger2012we,menze2015object,mayer2016large}, deep learning-based algorithms with much improved performance have been developed. 
After \etal{Zagoruyko} \cite{zagoruyko2015learning} propose a deep convolutional neural networks for patch comparison,
the MC-CNN by Zbontar and LeCun \cite{zbontar2016stereo} is trained for stereo matching cost computation. 
While \cite{zbontar2016stereo} uses conventional cost aggregation methods, \etal{Kendal} \cite{kendall2017end} propose an end-to-end network performing all steps in 3D convolutional layers.

Our network is the first neural network-based stereo algorithm that learns omnidirectional cost maps from spherical input images.
As shown in the experiments it generates much cleaner cost volumes compared to the conventional intensity-based costs and covers 360\degree\ at once.
%
%
\begin{figure}
\centering
  \begin{subfigure}[b]{0.57\linewidth}
  \captionsetup{justification=centering}
  \includegraphics[width=\linewidth]{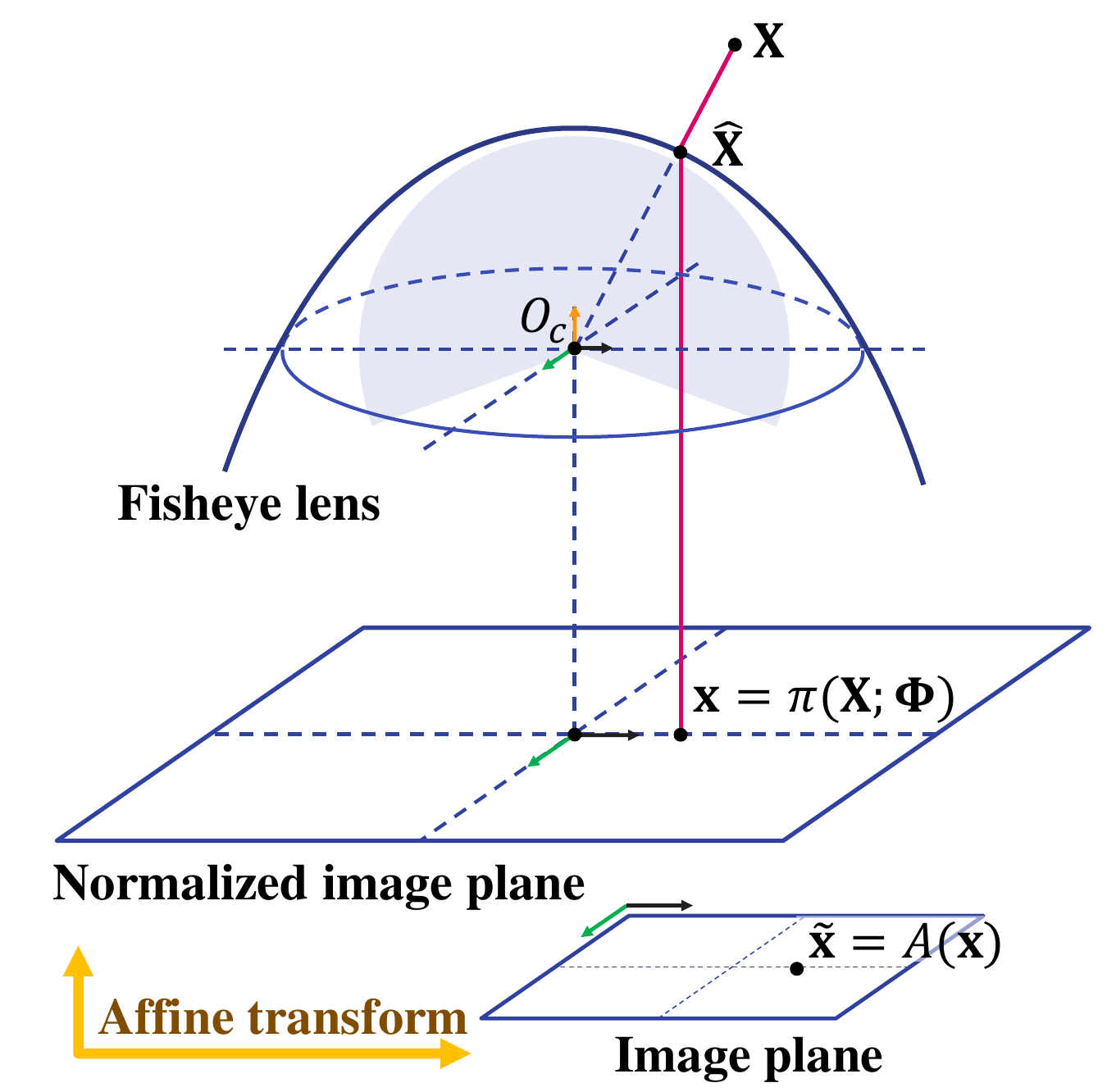}
  \caption{}\label{fig:projection}
  \end{subfigure}
  \begin{subfigure}[b]{0.38\linewidth}
  \includegraphics[width=\linewidth]{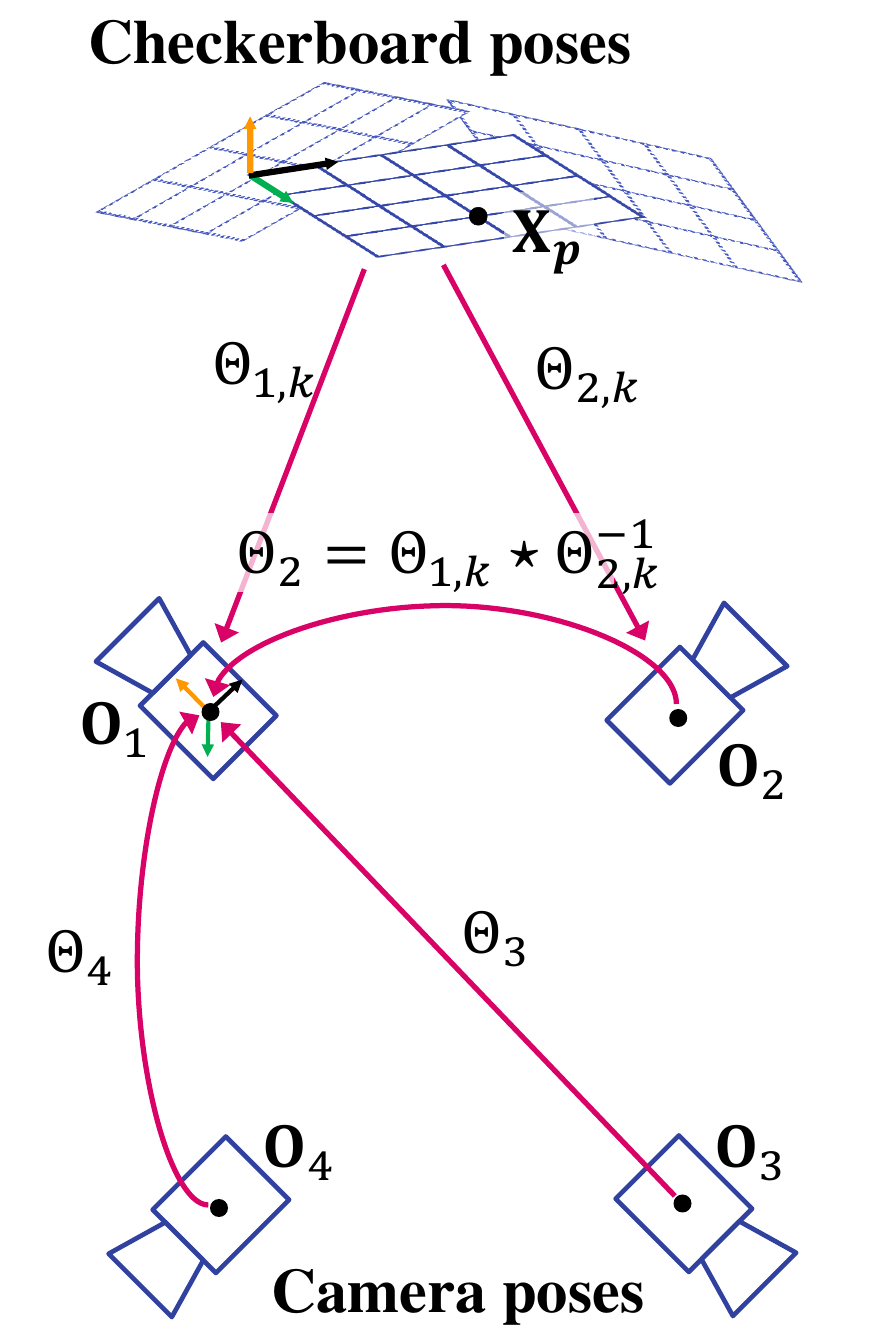}
  \captionsetup{justification=centering}
  \caption{}\label{fig:calibration}
  \end{subfigure}
  \caption{
(a) We use flipped catadioptric coordinate system for our fisheye lens \cite{scaramuzza2006flexible}. (b) We calibrate all the extrinsic parameters in the first camera's coordinate system.}
  \label{fig:calib}
  \vspace{-10pt}
\end{figure}

\section{OMNIDIRECTONAL STEREO}
\label{sec:omnidirectional}

\subsection{Fisheye Projection Model and Extrinsic Calibration}
\label{sec:extrinsic}

We use the omnidirectional camera model \cite{scaramuzza2006flexible, urban2015improved}, which models the lens distortion with a polynomial function. 
%
The projection function $\Pi$ maps a 3D point $\mathbf{X}$ to a 2D point $ \mathbf{x}$ on the normalized image plane, $\mathbf{x}=\Pi(\mathbf{X};\mathbf{\Phi}) $, where $\mathbf{\Phi}$ is the fisheye intrinsic parameters. 
%
%
The normalized image coordinate $\mathbf{x}$ is transformed to the pixel coordinate by an affine transformation $A(\mathbf{x})$, as in Fig.~\ref{fig:projection}.
%
%
The details of the projection models are described in \cite{scaramuzza2006flexible}.

We follow the conventional camera rig calibration procedure using a checkerboard - for each camera the lens intrinsic parameters and relative poses of the checkerboards are computed, then the rig is initialized using the relative poses, and finally all extrinsic and intrinsic parameters are optimized.
A large checkerboard is used to ensure sufficient overlaps between views.
The extrinsic parameters are represented as $\Theta = (\mathbf{r}^\top, \mathbf{t}^\top)^\top$ , where $\mathbf{r}$ is an axis-angle rotation vector and $\mathbf{t}$ is a translation vector ($\mathbf{r}, \mathbf{t} \in \mathbb{R}^3$).
The rigid transformation matrix $M(\Theta)$ is given as $\begin{bmatrix} R(\mathbf{r}) & \mathbf{t}\end{bmatrix}$ where $R(\mathbf{r})$ is the $3\times 3$ rotation matrix corresponding to $\mathbf{r}$.
From the checkerboard images of a camera $i$, we denote its lens intrinsics $\mathbf{\Phi}_i$ and $A_i$, as well as the checkerboard poses to the camera, $\{\Theta_{i,k}\}$ where $k$ is the capture index.
The relative pose from camera $i$ to $j$ can be computed as $\Theta_{j,k} \star \Theta_{i,k}^{-1}$ from a pair of simultaneously-taken images $(i,k)$ and $(j,k)$, where $\star$ and $^{-1}$ denotes the composition and inverse operations.
For extrinsic calibration, all camera poses $\{\Theta_i\}$ and the checkerboard poses $\{\Theta_k\}$ are initialized in the first camera's coordinate system, Fig.~\ref{fig:calibration}, and we minimize the reprojection error of the corner points on the checkerboards 
\[ \min_{\substack{\mathbf{\Phi}_i,A_i \\ \Theta_i,\Theta_k}} \sum_{(i,k)}\sum_{p} \left\|\: \tilde{\mathbf{x}}_{i,p} - A_i\!\left(\!\Pi\!\left(M(\Theta_i \star\Theta_k) \!\begin{bmatrix} \mathbf{X}_p \\ 1\end{bmatrix}\! ; \mathbf{\Phi}_i \right)\!\!\right) \right\|^2 \!,\]
where $\{(i,k)\}$ is the set of observations of the checkerboard pose $k$ with the camera $i$, $\mathbf{X}_p$ is the coordinate of the corner point $p$ in the checkerboard, and $\tilde{\mathbf{x}}_{i,p}$ is the pixel coordinate of $\mathbf{X}_p$ in the image $i$.
Ceres solver~\cite{ceres-solver} is used in optimization.

\begin{figure}
\centering
  \includegraphics[width=0.95\linewidth]{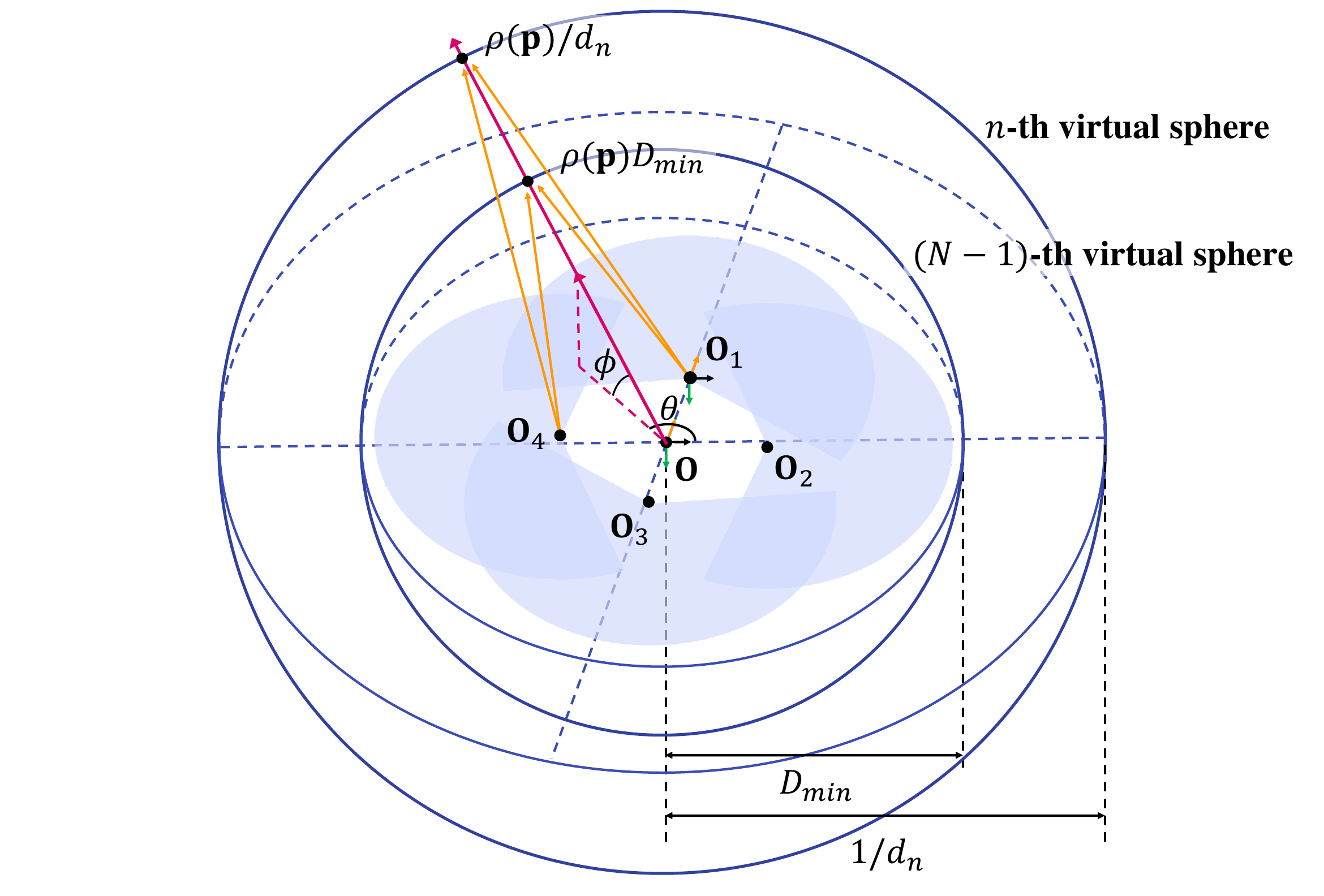}
  \caption{
Warping the input images onto each virtual sphere, we measure the stereo matching costs from the pairwise images.}
  \label{fig:omni}
  \vspace{-10pt}
\end{figure}
\begin{figure*}[t]
\centering
    \includegraphics[keepaspectratio=true,width=0.97\textwidth]{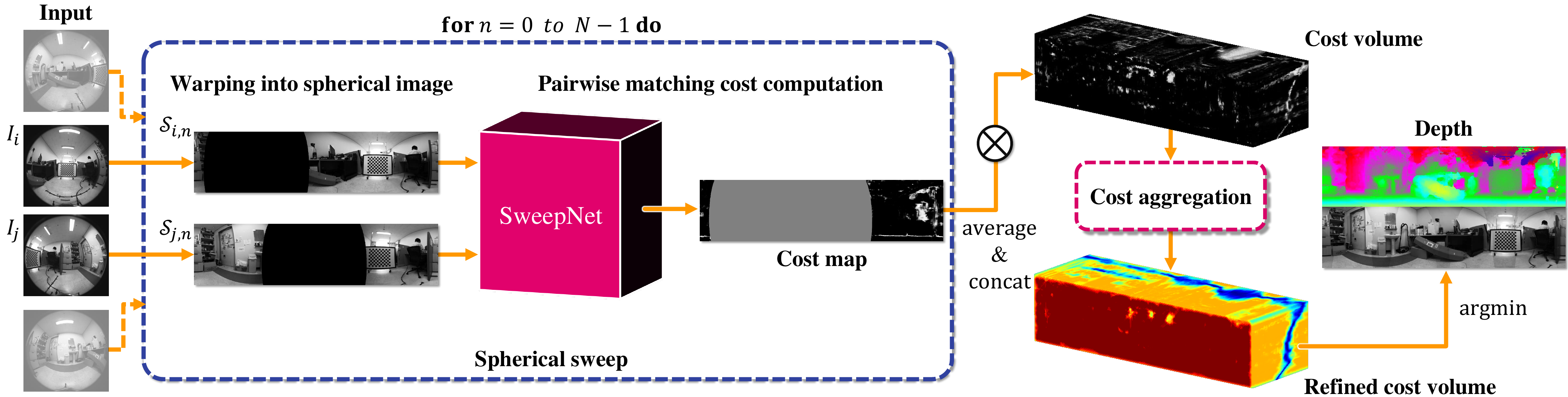}
    \caption{
    Flowchart of the proposed method.
    We warp the selected pair of images onto the $n$-th sphere. SweepNet computes the matching cost from the input spherical image pair $\mathcal{F}(\mathcal{S}_{i,n}, \mathcal{S}_{j,n})$, then we build the cost volume $\mathcal{C}$ by average of the costs of all the selected pairs. The omnidirectional depth can be acquired through cost aggregation and winner-takes-all strategy.}
    \label{fig:workflow}
    \vspace{-15pt}
\end{figure*}

\subsection{Spherical Sweep}
\label{sec:spherical}

The plane-sweep algorithm \cite{collins1996space} enables dense stereo matching among multi-view images. 
However, it is difficult to apply the algorithm to fisheye images with more than 180\degree\ FOV. 
\etal{Hane} \cite{hane2014real} use multiple planes with different normals and distances, and later \etal{Im} \cite{im2016all} exploit local spheres centered at the reference camera to estimate depth from a spherical panoramic camera.

The proposed system works with both wide FOV images in a wide-baseline setup, which cannot be handled by the existing algorithms.
To estimate omnidirectional depth in our wide-baseline system, we propose a global spherical sweep algorithm. 
The center of sweeps can be anywhere, but to minimize the distortion on the spherical images, we choose the rig center for the origin and align the $xz$-plane to be close to the camera centers.
%
%
In this spherical coordinate system, a ray $\mathbf{p} = (\theta,\phi)$ corresponds to  
$\rho(\mathbf{p})=(\cos(\phi)\cos(\theta),\, \sin(\phi),\, \cos(\phi)\sin(\theta))^\top$.
Let the transformed camera extrinsic parameters in the rig coordinate system be $\{\Theta^*_i\}$.
Also for notational simplicity, we denote the projection function $A_i(\Pi(\mathbf{X};\mathbf{\Phi}_i))$ with $\Pi_i(\mathbf{X})$.

We now warp the input images onto the global spheres. 
Each pixel in the warped spherical image $\mathcal{S}$ represents a ray $(\theta,\phi)$.
The spherical image $\mathcal{S}$ has $W\times H$ resolution and $\theta$ varies from $-\pi$ to $\pi$.
$\phi$ can be up to $-\pi/2$ to $\pi/2$, but we use a smaller range in our experiments as the ceiling (sky) and ground are of less interest.
$N$ spheres are sampled so that their inverse depths are uniform, i.e., when the minimum depth is $D_{min}$, the inverse depth to the $n$-th sphere is $d_n = n / (D_{min}(N-1))$, $n\in [0,...,N-1]$.
In other words, the radii of the spheres are $1/d_n$ except $n=0$, which corresponds to the sphere at infinity.
As shown in Fig.~\ref{fig:omni} the pixel value of the spherical image is determined as
\begin{equation}\label{eq:spherical}
\mathcal{S}_{i,n}(\mathbf{p}) = I_i\!\left( \Pi_i\!\left( M(\Theta^*_i)\begin{bmatrix} \rho(\mathbf{p})/d_n \\ 1\end{bmatrix}\right)\!\right),
\end{equation}
where $I_i$ is the input image captured by camera $i$.
For $n=0$, we use $d_0 = 2^{-23}$.
%
When the projected pixels are not in the visible region of the input image, we do not consider them in the further processing.

In the spherical sweep algorithms, we need to compute the $W \times H\times N$ matching cost volume $\mathcal{C}$ for all ray directions and inverse depths.
%
%
Suppose that we are given a pairwise matching cost function $\mathcal{F}(\cdot,\cdot)$ which takes two images and computes the cost map of the same size.
The integrated cost map is the average of all possible (and valid) pairwise cost maps, and the cost volume is the collection of integrated cost maps, i.e., the cost of $\mathbf{p}$ at $n$-th sphere is 
\begin{equation}\label{eq:costvolume}
\mathcal{C}(\mathbf{p},n)=\operatorname*{mean}_{ij} \Big{\{}\mathcal{F}(\mathcal{S}_{i,n}(\mathbf{p}),\mathcal{S}_{j,n}(\mathbf{p}))\Big{\}}
\end{equation}
where $ij$ is an unordered index pair of spherical images. 
As the raw cost volume is often noisy and contains incorrect estimates, we take advantage of SGM \cite{hirschmuller2008stereo} which refines the cost volume by performing minimization of an energy function with dynamic programming.
Finally, the inverse depth of a ray $\mathbf{p}$ is determined by the winner-takes-all strategy as $d_{n^*}$, where $n^* = \operatorname*{argmin}_n \mathcal{C}(\mathbf{p},n)$.
The overall procedure is illustrated in Fig.~\ref{fig:workflow}.

As a baseline cost function, we use zero mean normalized cross correlation (ZNCC), which is the covariance of two patches divided by their individual standard deviations.
ZNCC is one of the most popular cost functions \cite{faugeras1993real,gallup2007real}, since it is robust to radiometric changes.
However in our challenging setup, it does not generate good cost maps, thus we propose our neural network cost function.

\begin{table}
\vspace{5pt}
\centering
\begin{tabular}{@{}llc@{}}
 Layer & Property & Output Dim. \\ \midrule \midrule
input & add circular column padding & $ (W+4) \times H $ \\ \midrule
conv1 & $5 \times 5$, 32, s 2, $\textrm{p}_W$ 0, $\textrm{p}_H$ 2
& \multirow{5}{*}{$ \frac{1}{2}W \times \frac{1}{2}H \times 32 $} \\
conv2 & $3 \times 3$, 32, s 1, p 1 &  \\
conv3 & $3 \times 3$, 32, s 1, p 1, add conv1 &  \\
conv4-17 & repeat conv2-3 & \\
conv18 & $3 \times 3$, 32, s 1, p 1 &  \\ \midrule
concat &  & $ \frac{1}{2}W \times \frac{1}{2}H \times 64 $ \\ \midrule
conv19 & $3 \times 3$, 128, s 1, p 1 & $ \frac{1}{2}W \times \frac{1}{2}H \times 128 $ \\
deconv1 & $3 \times 3$, 128, s 2, p 1 & $ W \times H \times 128 $ \\
conv20 & $3 \times 3$, 128, s 1, p 1 & $ W \times H \times 128 $ \\
fc1-4 & $1 \times 1$, 256 & $ W \times H \times 256 $ \\
fc5 & $1 \times 1$, 1, no ReLu & $ W \times H $ \\
\midrule
sigmoid &  & $ W \times H $
\end{tabular}
\caption{SweepNet has 20 convolutional layers and a transposed convolutional layer followed by 5 fully connected layers. Each properties (s, p) means (stride, padding)  in the convolutional block. }
\label{table:sweepnet}
\vspace{-10pt}
\end{table}

\begin{table*}[hbt!]
\centering
\begin{threeparttable}
\begin{tabular}{l|rrrrr|rrrrr|rrrrr}
\multirow{2}{*}{Matching cost func.}& \multicolumn{5}{c|}{Sunny} & \multicolumn{5}{c|}{Cloudy} & \multicolumn{5}{c}{Sunset} \\ 
 & \multicolumn{1}{c}{\textgreater{}1} & \multicolumn{1}{c}{\textgreater{}3} & \multicolumn{1}{c}{\textgreater{}5} & \multicolumn{1}{c}{MAE} & \multicolumn{1}{c|}{RMS} & \multicolumn{1}{c}{\textgreater{}1} & \multicolumn{1}{c}{\textgreater{}3} & \multicolumn{1}{c}{\textgreater{}5} & \multicolumn{1}{c}{MAE} & \multicolumn{1}{c|}{RMS} & \multicolumn{1}{c}{\textgreater{}1} & \multicolumn{1}{c}{\textgreater{}3} & \multicolumn{1}{c}{\textgreater{}5} & \multicolumn{1}{c}{MAE} & \multicolumn{1}{c}{RMS} \\ \hline \hline
ZNCC & 40.7 & 28.0 & 25.2& 10.0 & 23.0 & 44.9 & 31.0 & 27.9 & 10.9 & 23.9 & 39.5 & 26.8 & 24.0 & 9.7 & 22.9 \\
MC-CNN \cite{zbontar2016stereo} & 42.1 & 32.7 & 30.2 & 13.3 & 27.8 & 46.1 & 33.5 & 30.6 & 12.8 & 26.5 & 42.9 & 33.0 & 30.5 & 13.8 & 28.4 \\
\textbf{SweepNet} & \textbf{20.7} & \textbf{13.2} & \textbf{11.2} & \textbf{4.1} & \textbf{14.0} & \textbf{28.5} & \textbf{16.4} & \textbf{14.0} & \textbf{5.0} & \textbf{15.0} & \textbf{20.5} & \textbf{13.6} & \textbf{11.7} & \textbf{4.6} & \textbf{15.6} \\ \hline
ZNCC + SGM \cite{hirschmuller2008stereo} & 24.0 & 9.9 & 6.3 & 1.5 & 4.5 & 25.6 & 9.9 & 6.3 & 1.6 & 4.5 & 23.4 & 9.9 & 6.4 & 1.6 & 4.6 \\
MC-CNN + SGM & 19.3 & 7.6 & 5.1 & 1.4 & 4.5 & 21.2 & 7.2 & \textbf{4.7} & 1.4 & 4.4 & 18.8 & 7.6 & 5.1 & 1.4 & 4.6 \\
\textbf{SweepNet} + SGM & \textbf{15.4} & \textbf{6.8} & \textbf{4.8} & \textbf{1.1} & \textbf{3.8} & \textbf{19.6} & \textbf{7.2} & 4.9 & \textbf{1.2} & \textbf{3.8} & \textbf{14.8} & \textbf{7.0} & \textbf{4.9} & \textbf{1.2} & \textbf{3.9}
\end{tabular}
\caption{Quantitative results on the synthetic datasets. The error function is defined in (\ref{eq:error}), and the qualifier '\textgreater{}$n$' refers to the ratio of pixels (\%) whose error is larger than $n$, 'MAE' refers to the mean absolute error, and 'RMS' refers to the root mean square of $e$. The errors are averaged over all 300 frames of the test sets. The SGM smoothness penalties are set to $P_1=0.1$ and $P_2=12$.}
\label{table:quantitative}
\end{threeparttable}
\vspace{-10pt}
\end{table*}

\subsection{SweepNet} \label{sec:sweepnet}

%
Local patch-based approaches including deep learning-based MC-CNN \cite{zbontar2016stereo} fails to find hard negative samples such as two identical patches from different objects since they do not consider holistic visual information.
Moreover, in the wide-baseline setting, the same patches may look different due to foreshortening or radiometric differences from viewing direction changes.
%
To handle this, we propose SweepNet which utilizes the global context in the images.

The architecture of the proposed network is detailed in Table~\ref{table:sweepnet}. 
As shown in Fig.~\ref{fig:workflow}, the input of the network is a pair of gray scale spherical images acquired from (\ref{eq:spherical}). 
To ensure that the horizontal ends are connected, we add the circular column padding to the input spherical images. 
The conv1$\sim$18 layers are Siamese residual blocks \cite{he2016deep} for learning the unary feature extraction. 
We reduce the size of the input image in half for the larger receptive field, which helps the network learns from global context. 
The output feature maps are concatenated, and then the features are upsampled using transposed convolution. 
Finally, the network outputs the $W \times H$ cost map which ranges from 0 to 1, through fully connected layers and a sigmoid layer.

To train our network, we use the following approach.
Given a set of $W \times H$ ground-truth depth maps $\{\hat{\mathcal{D}}_l\}$, the inverse depth index is given as $\hat{n}_l(\mathbf{p}) = \operatorname*{round}(D_{min}(N-1)/\hat{\mathcal{D}}_l(\mathbf{p}))$.
Each position $\mathbf{p}$ on $n$-th sphere is labeled as
\newcommand\round[1]{\left[#1\right]}
\begin{equation}\label{eq:gt}
\hat{y}(l, \mathbf{p}, n)=\begin{cases}0 \,(positive), &\text{if } \hat{n}_l(\mathbf{p}) = n \\ 1 \,(negative), & \text{otherwise}\end{cases}
\end{equation}
We use the negative binary cross-entropy loss~\cite{robert2014machine}, $\Lambda(\hat{v},v) = -(\hat{v}\log v + (1-\hat{v})\log(1-v))$.
For the labeled training set $L = \{\hat{y}(l,\mathbf{p},n)\}$, the loss is defined as
\begin{equation}\label{eq:loss}
\mathcal{L}(y)=\frac{1}{|L|}\sum_{(l,\mathbf{p},n)\in L} \Lambda(\hat{y}(l,\mathbf{p}, n), y(l,\mathbf{p}, n))
\end{equation}
where $y(l,\mathbf{p}, n)$ is the predicted label at $(\mathbf{p},n)$ with the input images corresponding to $\hat{\mathcal{D}}_l$.
The loss is minimized by stochastic gradient descent with a momentum.  

\section{EXPERIMENTAL RESULTS}
\label{sec:experimental}

\label{sec:system}

\subsec{System Configuration}
We use four CCD cameras with 220\degree\ FOV fisheye lenses (Pointgrey CM3-U3-31S4C and Entaniya M12-220).
$4\times (1600\times 1532)$ images can be captured at 30 Hz, and they are synchronized by software trigger.
For indoor experiments, we use a square-shaped rig ($300\times 300$ mm), and for outdoors, the cameras are installed at the four corners of the roof of a minivan as shown in Fig.~\ref{fig:intro}.
For calibration, a checkerboard with $12\times 10$ grids each of which is $60\times 60$ mm is used.



\begin{figure}
\vspace{-5pt}
\centering
\includegraphics[width=0.95\linewidth]{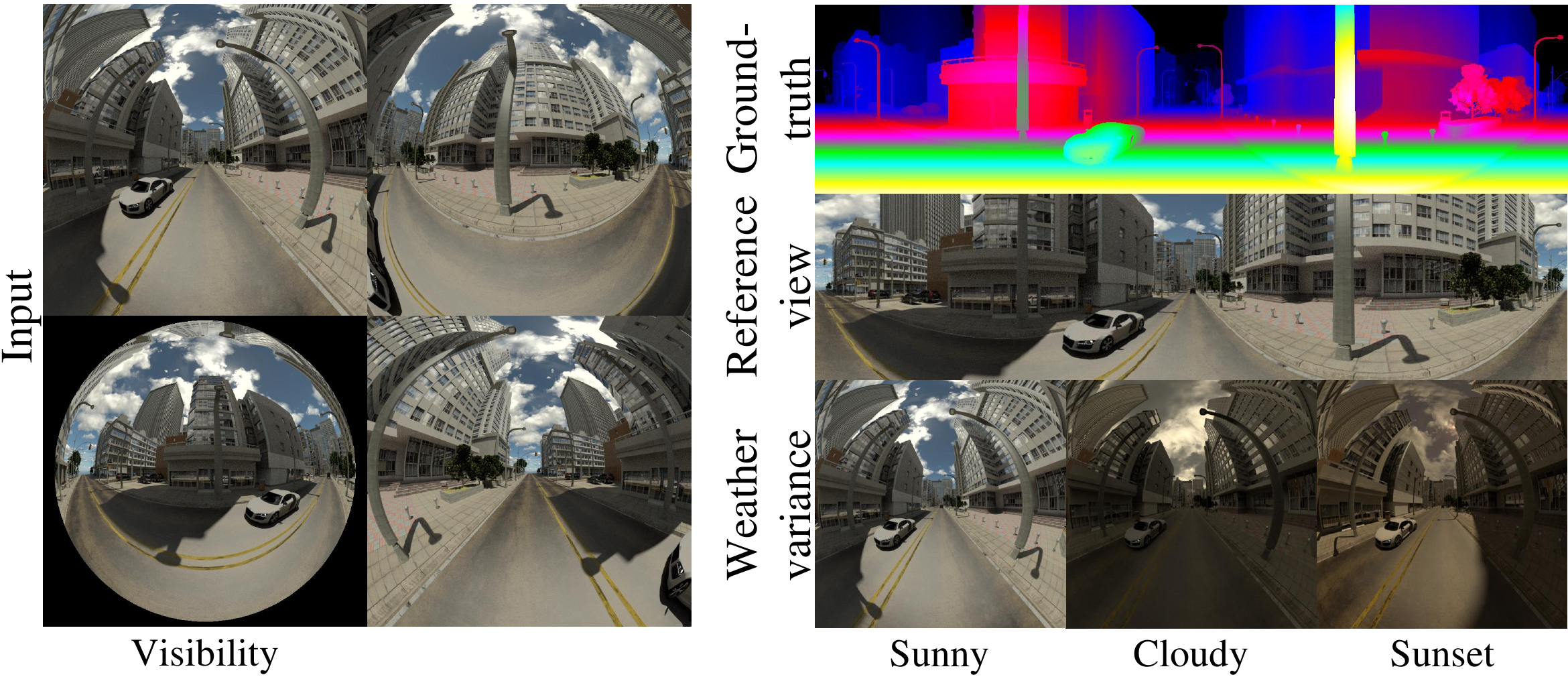}
\caption{We construct the virtual omnidirectional system. As for the input images, we only use pixels within 220\degree\ FOV. }
\label{fig:datasets}
\vspace{-15pt}
\end{figure}

\subsec{Training}
In order to train the network, we create the synthetic urban datasets with Blender as shown in Fig.~\ref{fig:datasets}.
Following \cite{zhang2016benefit}, we virtually implement the camera rig similar to our outdoor setting as well as buildings, cars, and roads in Blender, and render each frame as four $800\times 768$ images.
The Sunny dataset consists of 1000 sequential frames of sunny city landscapes, and we split them into two parts, the former 700 frames for training and the later 300 for testing. 
We also create separate test datasets with varying weather (Cloudy, and Sunset) to test on different photometric conditions.
%

%
The input images are converted to grayscale, the intensity values are normalized to zero-mean and unit variance, and they are warped to the spherical images of $W=600$ and $H=150$ for training.
We set $\phi$ from $-45\degree$ to $45\degree$ for the synthetic datasets.
%
Among the training data, we randomly select 350 frames and train our network for 14 epochs. 
The learning rate is initially set to $0.003$ for the first 11 epochs and $0.0003$ for the remaining.
%
%
We sample 192 inverse depths, and the corresponding ground-truth labels are acquired by (\ref{eq:gt}).
The inverse depths with less than $32$ positive labels are discarded, and the same number of positive and negative labels are used for training.
%
In total 92 million labels from the Sunny training dataset are used to train our network.

\begin{figure}[hbt!]
\centering
  \begin{subfigure}[b]{\linewidth}
  \captionsetup{justification=centering}
  \includegraphics[width=\linewidth]{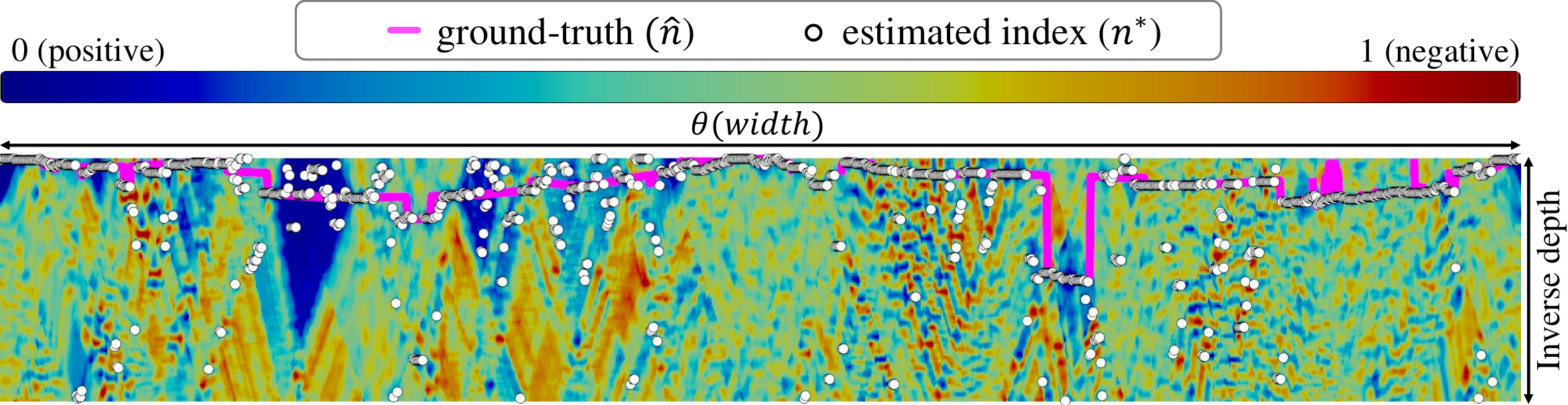}
  \caption{ZNCC}\label{fig:cross_section_zncc}
  \end{subfigure}
  \begin{subfigure}[b]{\linewidth}
  \captionsetup{justification=centering}
  \includegraphics[width=\linewidth]{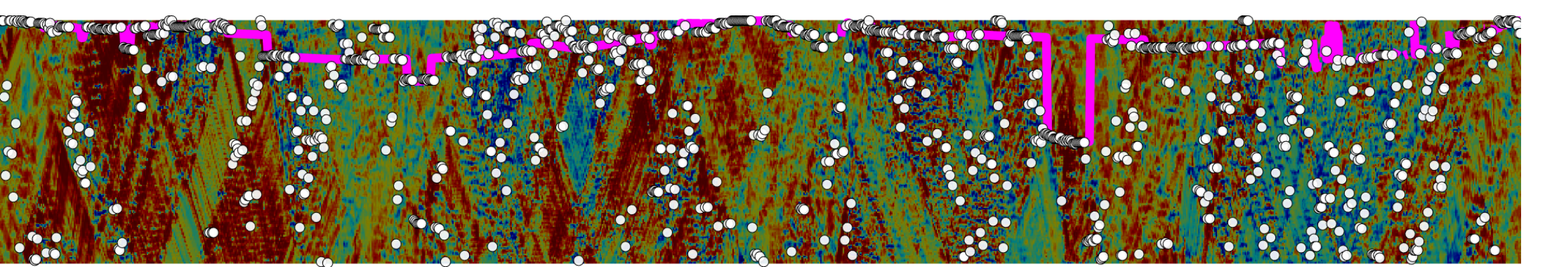}
  \caption{MC-CNN}\label{fig:cross_section_mccnn}
  \end{subfigure}
  \begin{subfigure}[b]{\linewidth}
  \captionsetup{justification=centering}
  \includegraphics[width=\linewidth]{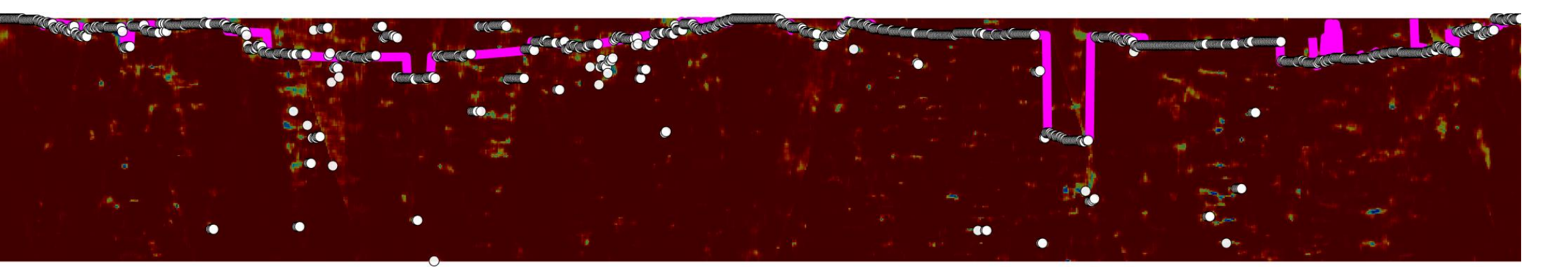}
  \caption{\textbf{SweepNet}}\label{fig:cross_section_sweepnet}
  \end{subfigure}
  \caption{Cross section of the raw cost volume cut along the $xz$-plane ($\phi=0\degree$). SweepNet classifies negative points more precisely.}
  \label{fig:cross_section}
  \vspace{-15pt}
\end{figure}

\begin{figure*}[hbt!]
	\centering
    \includegraphics[width=0.97\textwidth]{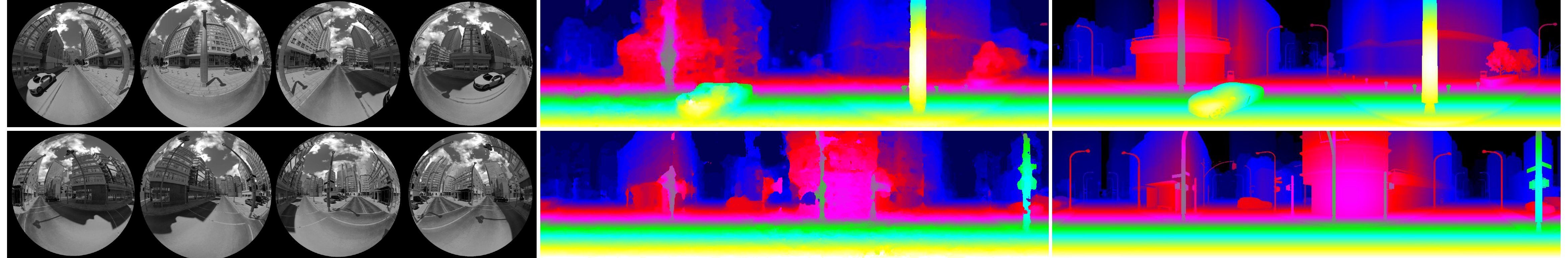}
    \caption{Qualitative results on the Sunny dataset. From left: input images, inverse depth prediction, and ground-truth inverse depth.}
    \label{fig:qualitative_synthetic}
\end{figure*}

\begin{figure*}[hbt!]
\vspace{-3pt}
	\centering
	\begin{subfigure}{0.97\textwidth}
    \captionsetup{justification=centering}
    \includegraphics[width=\textwidth]{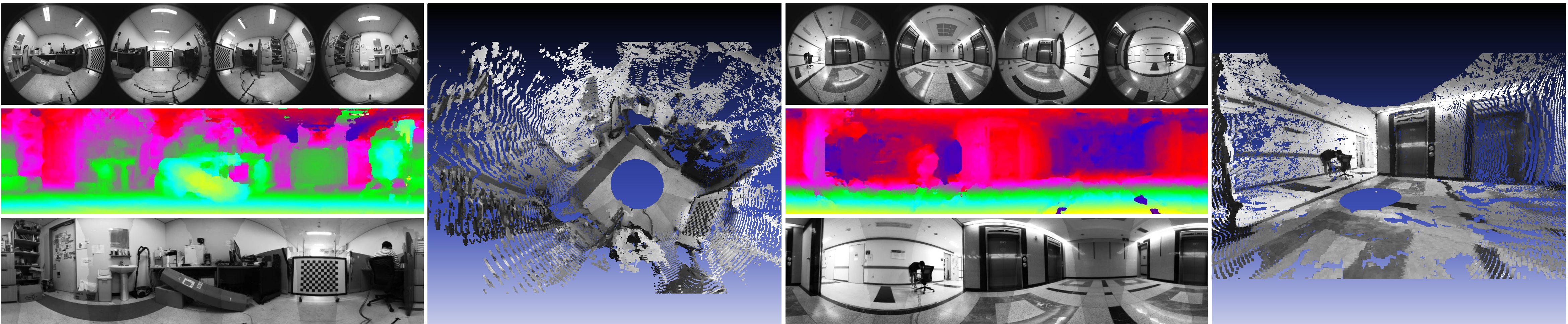}
    \end{subfigure}
    \begin{subfigure}{0.97\textwidth}    \captionsetup{justification=centering}
    \includegraphics[width=\textwidth]{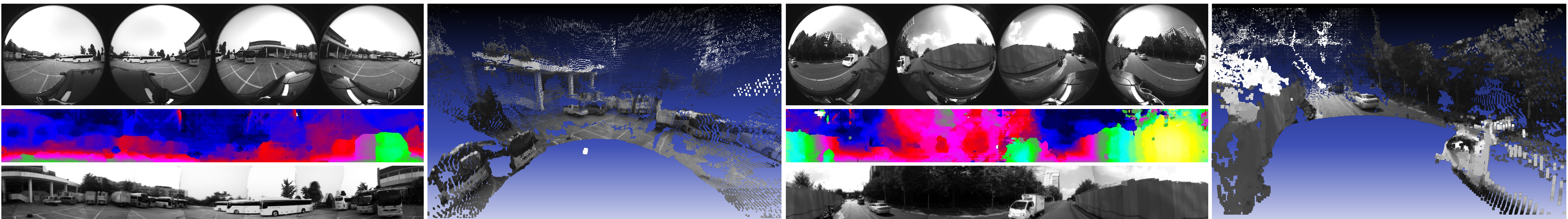}
    \end{subfigure}
    
    \caption{Qualitative results on the real world data. Left: input images, inverse depth prediction, and reference panorama image. Right: point cloud result.
    We set the size of the spherical image to $W=1200$ and $H=300$ and $\phi$ ranges from $-45\degree$ to $45\degree$ for indoor data. On the other hand, for outdoor data, we set the height of the spherical image to $H=150$ and $\phi$ ranges from $-15\degree$ to $30\degree$. We set $N=192$, $P_1=0.1$ and $P_2=12$ for both datasets. We use the network trained by the Sunny dataset.}
    \label{fig:qualitative_real}
    \vspace{-10pt}
\end{figure*}

\subsec{Evaluation}
We evaluate our method quantitatively on the synthetic datasets.
The error of inverse depth index is defined as 
\begin{equation}
\label{eq:error}
e(\mathbf{p}) = \frac{100}{N}|n^*(\mathbf{p})-\hat{n}(\mathbf{p})|
\end{equation}
The number of inverse depth $N$ is set to $192$ and the size of spherical image is $W=1200$ and $H=300$ in testing.

We compare the proposed SweepNet with other matching cost functions, ZNCC and MC-CNN \cite{zbontar2016stereo} on the Sunny, Cloudy and Sunset test sets.
%
The ZNCC window size for local patches is set to $9\times9$, and as high ZNCC values mean same patches, we use the negative ZNCC cost by $(1-\mathcal{F}_{ZNCC})/2$ which ranges from 0 to 1.
We train MC-CNN on the Sunny dataset by $40$ million pairs of $9\times9$ local patches from the spherical images with 14 epochs and 256 batches following the original literature \cite{zbontar2016stereo}. 
We compare the accuracy of depth maps with and without cost volume refinement by SGM~\cite{hirschmuller2008stereo}.
Table~\ref{table:quantitative} shows that the SweepNet outperforms other methods in all metrics. 
Especially, the SweepNet with SGM gives the best and most robust results in all datasets.
%
%

Fig.~\ref{fig:cross_section} shows the cross section of the raw cost volume at $\phi=0$ of the matching cost functions (without SGM).
The cost maps by ZNCC and MC-CNN have lots of false positives (green to blue colors outside the ground-truth depths), where SweepNet generates a much cleaner cost map.
%

\begin{figure}[ht!]
\centering
\includegraphics[width=0.9\linewidth]{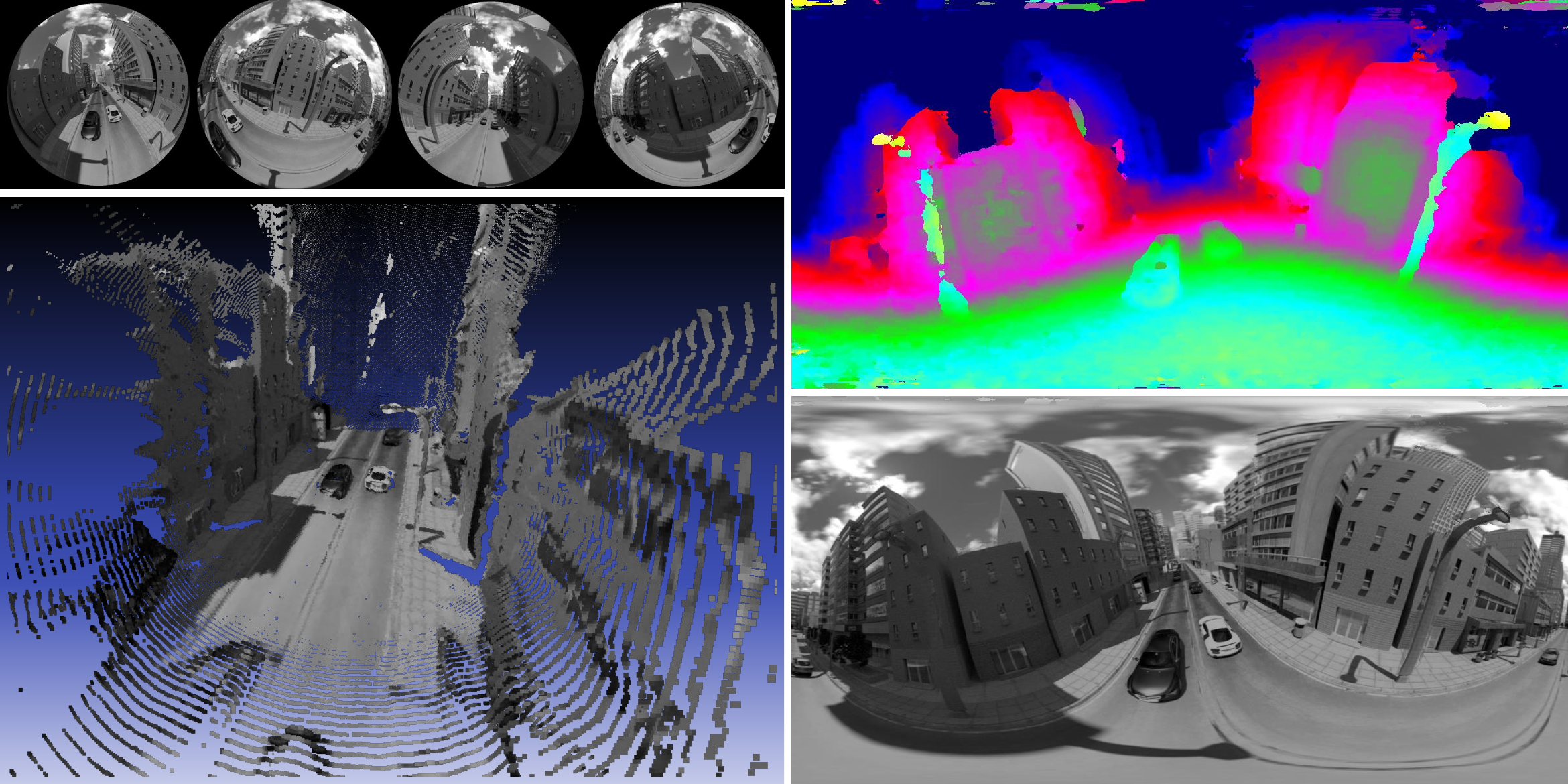}
\caption{Qualitative results on an aerial situation. The height of the spherical image is $H=600$, where $\phi$ ranges from $-90\degree$ to $90\degree$.}
\label{fig:aerial_result}
\vspace{-10pt}
\end{figure}

In Fig.~\ref{fig:qualitative_synthetic}, the estimated inverse depth map is shown with the ground-truth.
The buildings, car, and thin structures like traffic signs and poles are reconstructed successfully, and the sky and ground plane are also accurately estimated.
In wide-baseline, the thin structures are especially challenging, because in the cost volume, there can be multiple true matches for one ray direction, one at the thin object and another at the object behind it.
Fig.~\ref{fig:aerial_result} shows an example of full spherical depth estimation ($\phi$ varies from $-\pi/2$ to $\pi/2$), which is useful for drones that can move freely in all 6-DOF.
One can verify that the scene including sky and ground are precisely reconstructed, even when the $xz$-plane is not aligned with the ground plane.
%
%

In addition, we qualitatively evaluate the proposed method with the real world data captured by our indoor and outdoor rigs.
%
Fig.~\ref{fig:qualitative_real} shows the input images, the omnidirectional inverse depth map, the reprojected panorama image, and the 3D rendering of the point cloud.

In the indoor examples, one can see that the regions very close to the rig (such as the floor) are accurately reconstructed, and the walls with little texture are also well recovered.
The outdoor scenes are quite challenging since the far objects appear very small in the input images, whereas the near objects cover significant portions of the view.
The estimated inverse depth maps show that SweepNet can reconstruct both far and near objects successfully.
%
%
The panorama images at the bottom-left corners are constructed by projecting the estimated 3D points to the input images, which show how accurate the estimated depths are.
Aside the radiometric variations between cameras,
it is hard to find any mismatches in the panorama images.
The experiments
show that the proposed method can effectively handle the wide-baseline omnidirectional depth estimation problem.

\section{CONCLUSIONS}

We present a novel hardware system and stereo algorithm for omnidirectional depth estimation.
The proposed hardware configuration includes multiple widely placed cameras with wide FOV fisheye lenses.
After the intrinsic and extrinsic parameters are calibrated, the input images are warped into the spherical images by projection onto the virtual spheres positioned at the rig center with the predefined radii.
The proposed SweepNet considers holistic visual content in the spherical images at each radius to build the cost map.
%
%
%
With the proposed training data from the synthetically rendered city dataset, the SweepNet can be successfully trained.
The extensive experiments show that the SweepNet outperforms the local patch-based methods, and robustly generates accurate depth maps in challenging situations.
%
%


%

\section*{ACKNOWLEDGMENT}
\small{
This research was supported by Samsung Research Funding \& Incubation Center for Future Technology under Project Number SRFC-TC1603-05, Next-Generation Information Computing Development Program through National Research Foundation of Korea(NRF) funded by the Ministry of Science, ICT (NRF-2017M3C4A7069369), and the NRF grant funded by the Korea government(MISP)(NRF-2017R1A2B4011928).}


\bibliographystyle{IEEEtran}
\bibliography{IEEEexample}


\addtolength{\textheight}{-12cm}   

\end{document}